\documentclass[letterpaper]{article} 
\usepackage{aaai24}  
\usepackage{times}  
\usepackage{helvet}  
\usepackage{courier}  
\usepackage[hyphens]{url}  
\usepackage{graphicx} 
\usepackage{natbib}  
\usepackage{caption} 
\usepackage{algorithm}
\usepackage{algorithmic}

\usepackage{lipsum}
\usepackage{amsmath}
\usepackage{booktabs}
\usepackage{multirow}
\usepackage{amssymb}

\usepackage{amssymb}  
\usepackage{booktabs} 
\usepackage{multirow} 
\usepackage{array}    

\newcolumntype{R}[1]{>{\raggedleft\arraybackslash}p{#1}}
\newcommand{\specialcell}[2][c]{%
  \begin{tabular}[#1]{@{}c@{}}#2\end{tabular}}

\usepackage{newfloat}
\usepackage{listings}
\DeclareCaptionStyle{ruled}{labelfont=normalfont,labelsep=colon,strut=off} 
\floatstyle{ruled}
\newfloat{listing}{tb}{lst}{}
\floatname{listing}{Listing}
\title{GraphGuard: Contrastive Self-Supervised Learning \\ for Credit-Card Fraud Detection in Multi-Relational Dynamic Graphs}
\author {
    Kristófer Reynisson\textsuperscript{\rm 1},
    Marco Schreyer\textsuperscript{\rm 1},
    Damian Borth\textsuperscript{\rm 1}
}
\affiliations {
    \textsuperscript{\rm 1}University of St. Gallen (HSG), St. Gallen, Switzerland\\
    kristofer.reynisson@student.unisg.ch, marco.schreyer@unisg.ch, damian.borth@unisg.ch
}

\begin{document}

\maketitle

\begin{abstract}
Credit card fraud has significant implications at both an individual and societal level, making effective prevention essential. Current methods rely heavily on feature engineering and labelled information, both of which have significant limitations. In this work, we present GraphGuard, a novel contrastive self-supervised graph-based framework for detecting fraudulent credit card transactions. We conduct experiments on a real-world dataset and a synthetic dataset. Our results provide a promising initial direction for exploring the effectiveness of graph-based self-supervised approaches for credit card fraud detection.
\end{abstract}

\section{Introduction}

Credit card fraud poses a significant challenge, impacting victims both financially~\cite{uk_finance_uk_2022} and psychologically~\cite{button_not_2014}. Industry projections estimate that online payment fraud will result in global merchant losses totalling approximately \$343 billion between 2023 and 2027~\cite{maynard_fighting_2022}. Beyond individual repercussions, credit card fraud has broader societal implications, as it often finances illicit activities such as terrorism and drug trafficking~\cite{ryman-tubb_how_2018}. Additionally, the nature of fraud is evolving, with perpetrators increasingly employing more targeted social engineering schemes~\cite{uk_finance_uk_2022}. Consequently, advancing and adopting effective fraud detection strategies is critical to maintaining financial security and reinforcing consumer confidence in payment systems.

Training data-driven models for credit card fraud detection presents unique challenges. The \textit{class imbalance} between genuine and fraudulent transactions hinders traditional learning methods, making it difficult to effectively detect the minority class of fraudulent transactions~\cite{dal2015calibrating}. Additionally, the similar characteristics of fraudulent and genuine transactions, known as \textit{overlapping data}, necessitate analyzing transactions in the context of historical spending patterns~\cite{van_vlasselaer_apate_2015}. Another complexity is \textit{concept drift}, where consumer spending patterns and fraud strategies evolve over time, diminishing the effectiveness of static methods like rule-based systems and supervised learning~\cite{alippi2013just, dal_pozzolo_credit_2018}. Moreover, the dynamic nature of cardholder spending patterns means that classifiers need to be able to discriminate between normal and abnormal changes in cardholder spending in order to overcome the problem of overlapping data in the presence of concept drift~\cite{van_vlasselaer_apate_2015}. Finally, \textit{verification latency} delays feedback to \textit{fraud detection systems} (FDSs), limiting the use of the most recent, unlabeled data for training supervised learning models and exacerbating the challenges posed by concept drift~\cite{dal_pozzolo_credit_2018}.

\begin{figure*}[]
    \centering
    \includegraphics[width=\textwidth]{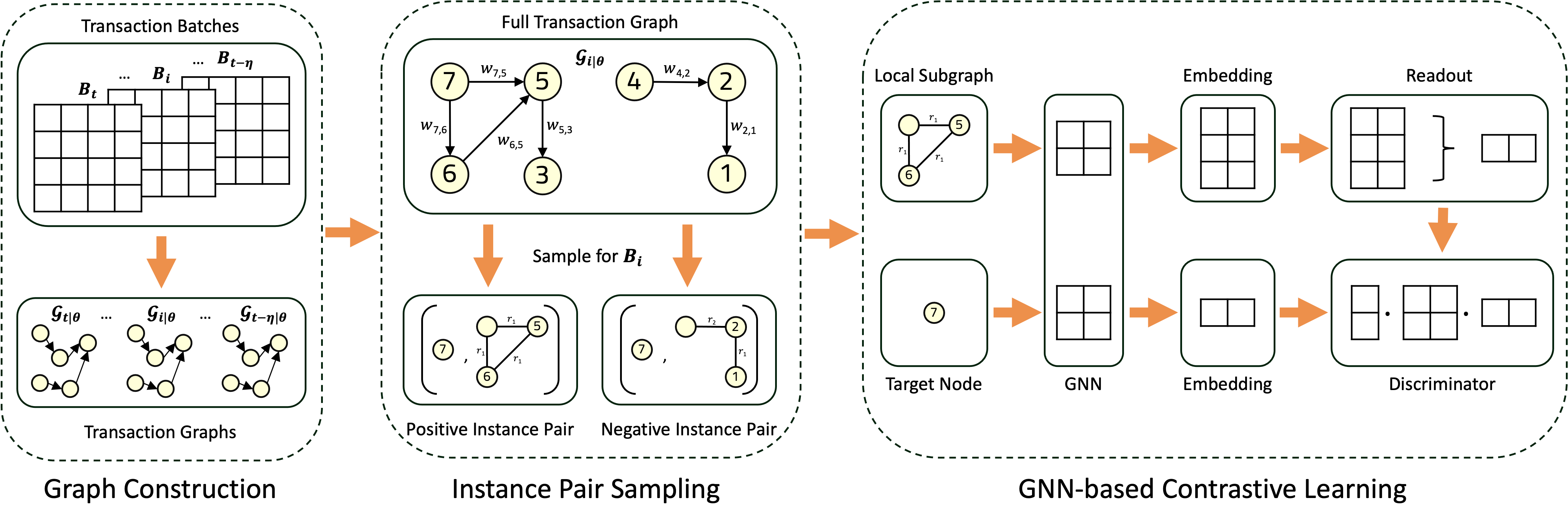}
    \caption[GraphGuard]{Schematic of GraphGuard, inspired by \cite{liu_anomaly_2022_COLA}. During \textit{Graph Construction}, each batch $B_i$ comprising daily transactions is transformed into a graph, incorporating data from a sliding window encompassing the most recent $\theta$ days. In the \textit{Instance Pair Sampling} phase, contrastive instance pairs are sampled for the transaction nodes of day $i$, utilizing preceding transactions as a historical backdrop. Lastly, \textit{GNN-Based Contrastive Learning} processes the target node and its local subgraph through a GNN to yield node embeddings. The node embeddings of the local subgraph are additionally put through a readout function to summarize their embeddings in an embedding vector. The embedding vector of the subgraph and the target node are then given to a discriminator module, which determines the fit between the target node and the local subgraph.}
    \label{fig:GraphGuard}
\end{figure*}

Addressing the challenge of overlapping data has traditionally relied on manual feature engineering, but these methods are limited in their ability to achieve optimal classification performance and cost-effectiveness~\cite{van_belle_catchm_2023}. Feature engineering methods are commonly designed to provide models with aggregated features to summarize cardholders' and merchants' recent activities~\cite{van_vlasselaer_apate_2015}. Thus, they serve to highlight transactions that are not anomalous by themselves but might be unusual compared to the previous spending patterns of the specific cardholder or merchant. Although these techniques have been successful in aiding fraud detection in a tabulated data format~\cite{dal_pozzolo_credit_2018, van_vlasselaer_apate_2015, paldino_role_2022}, they inherently discard the complex relationships between transactions~\cite{akoglu_graph_2015}. 

Many credit card fraud detection studies neglect the impact of verification latency, often assuming that labels are received regularly (e.g. the next day)~\cite{dal_pozzolo_credit_2018}. However, verification latency is crucial to consider, as it hampers classifiers' ability to adapt to new concepts, thus reducing their effectiveness~\cite{dal_pozzolo_credit_2018}. To counter this, some research has leveraged the alert-feedback interaction~\cite{lebichot_graph_semi_2017, dal_pozzolo_credit_2018}. Investigators review a small portion of alerted transactions arriving each day; thus, their true label is known, which has been used to provide a supervision signal during classifier training~\cite{dal_pozzolo_credit_2018}. While this method helps classifiers keep up with concept drift, it ignores the vast majority of unlabeled transactions in favour of a minority of labelled feedbacks and suffers from \textit{sample selection bias} as the selection of transactions into the set of feedbacks depends entirely on the alerts generated by the FDS~\cite{dal_pozzolo_credit_2018}. 

Deep learning has revolutionized tasks like object detection, traditionally reliant on handcrafted features \cite{ren_faster_2015}. Its success partly stems from exploiting the statistical properties of Euclidean data, where points in a multidimensional space enable the use of geometric principles vital for many machine learning algorithms. However, the application of deep learning to non-Euclidean data, such as graphs, is increasingly significant, particularly in modelling relationships in networks like social or citation networks. Unlike Euclidean data, graph data lacks standard properties like a global parameterization or shift-invariance, complicating operations like convolution~\cite{bronstein_geometric_2017}. This has spurred interest in \textit{graph neural networks} (GNNs), which preserve attribute and relational information in graph data and have been effective in fraud detection \cite{van_belle_catchm_2023}. In fraud detection, GNNs analyze credit card data as a graph, identifying fraudulent transactions by considering their attributes and the structure of related transactions, offering a solution to the challenges of overlapping data.

\textit{Self-supervised learning} (SSL) is recognized for its data efficiency and improved generalization, contrasting with the limitations of supervised learning \cite{liu_self_supervised_survey_2023}. SSL excels without labelled data, relevant for credit card fraud detection, addressing challenges like verification latency, concept drift, and class imbalance \cite{liu_SSL_imbalance_2022}. Combining SSL with deep graph representation learning offers a promising approach for credit card fraud detection, effectively analyzing data as a graph to identify fraudulent transactions by examining both transaction attributes and related transaction structures.

This work investigates self-supervised graph representation learning for credit card fraud detection. In summary, we present the following contributions.

\begin{itemize}

\item We enhance contemporary graph-based contrastive self-supervised learning anomaly detection frameworks by incorporating time-weighted subgraph sampling. 

\item We propose GraphGuard, a self-supervised graph representation learning framework for credit card fraud detection in dynamic multi-relational graph structures.

\item We evaluate our proposed framework employing real-world and synthetic datasets.

\end{itemize}

\section{Related Work}

This section covers two key areas related to our study: fraud detection in graphs and self-supervised methods for anomaly detection. We end the chapter by highlighting the specific area our research aims to explore and contribute to.

\subsection{Fraud Detection in Graphs}

This section focuses on work in credit card fraud detection. For a comprehensive review of graph-based anomaly detection approaches, readers should refer to the survey by~\citet{ma_comprehensive_2021}.

Early work employing graphs for credit card fraud detection was based on \textit{social network analysis}. \citet{van_vlasselaer_apate_2015} were among the first works to exploit the relational structure of credit card transactions within graphs to detect fraud. They used the labelled fraudulent transactions to propagate fraud signals from fraudulent transaction nodes throughout the network, allowing them to estimate the fraud exposure of new transactions. 

Graph representation learning has recently emerged in the credit card fraud detection literature as an alternative to feature engineering.~\citet{van_belle_inductive_2022} assesses the practicality of inductive graph representation learning algorithms for fraud detection in credit card transactions. This work is extended in~\citet{van_belle_catchm_2023}. Here, they integrate a supervised signal into the graph by introducing an artificial fraud node connected to all fraudulent transactions in the training set, ensuring that fraudulent nodes receive similar embeddings. Their findings show that graph representation learning approaches can outperform state-of-the-art feature engineering approaches. ~\citet{liu_gnn_ccfraud_2021} adopt a different approach using homogeneous transaction graphs. They additionally introduce a sampling policy and an attention mechanism.

\subsection{Self-Supervised Anomaly Detection}

The following section is built on the survey by~\citet{hojjati_self_supervised_anomaly_survey_2023} on self-supervised methods for anomaly detection but only reviews the graph-based methods. Readers are referred to the original survey for a review of other approaches. 

~\citet{liu_anomaly_2022_COLA} propose an approach based on contrasting nodes with their local subgraph. Building upon this concept,~\citet{zheng_generative_2021} use two pretext tasks: one for contrasting the target node against positive and negative samples and another for reconstructing the target node's features from positive subgraphs. Unlike the previous approaches,~\citet{zheng_unsupervised_2022} integrate data labels and explore few-shot anomaly detection scenarios where only a few known anomalies are available. These known labels provide a supervision signal to refine the training process and enhance the accuracy of anomaly detection. Recent studies have extended the scope of investigation from static graphs to dynamic graphs.~\citet{liu_anomaly_2021} address edge-level anomaly detection over time within dynamic graphs.

The literature review conducted for this work indicates that, to the best of our knowledge, this is the first work to leverage self-supervised learning of graph representations within the credit card fraud detection domain. 

\section{Problem Formulation}\label{sec:problem_formulation}

This section formalizes key credit card fraud detection components from the complex details inherent in real-world FDSs. The formulation presented here is based on formulations by \citet{dal_pozzolo_credit_2018} and \citet{paldino_role_2022}.

Let the vector $x_i$ denote the feature vector of the $i^{\mathrm{th}}$ transaction. Each $x_i$ is accompanied by a binary class label $y_i \in \{0, 1\}$, where 1 signifies a fraudulent transaction and 0 a legitimate one. In a real-world FDS, transactions arrive in a continuous stream \cite{carcillo_scarff_2017}. Inspired by \citet{paldino_role_2022}, the streaming nature of transaction data is simplified by modelling it in batches ${B}_t$ arriving daily. While a simplification, this approach accurately reflects the operational practice in FDSs, where classifiers are updated daily with the latest data and then applied to the next day of data \cite{dal_pozzolo_credit_2018}. The batch arriving at the end of day $t$ is denoted an ordered set $B_t = \{ x_j, x_{j+1}, ..., x_{j+s-1} \}$. The set is ordered based on transaction time, such that $j < j'$ implies $x_{j,\textrm{time}} < x_{j',\textrm{time}}$.

Moreover, to model the time-evolving characteristics of the transaction dataset effectively, let $X_{t|\eta} = \{ B_{t-\eta}, ..., B_{t-1} \}$ denote the set of transactions comprising the $\eta$ most recent batches available at time $t$.

The goal is to train a classifier $\mathcal{K}_{t|\eta}$ on the transactions in $X_{t|\eta}$ and to classify the transactions from day $t$, namely $B_t$. The dynamic nature of fraud means that classifiers are typically trained on the $\eta$ most recent days of available data where the value of $\eta$ balances the timeliness of the data, as too large a value might lead to the classifier learning outdated concepts, and the size of the training dataset, as the dataset needs to be large enough for the classifier to be effective. 

\section{Methodology}

This section details the methodology behind our proposed learning framework, GraphGuard. As shown in Figure~\ref{fig:GraphGuard}, the framework comprises three phases: Graph Construction, Instance Pair Sampling, and GNN-Based Contrastive Learning Model. Following is a description of each phase. 

\subsection{Graph Construction}\label{sec:methodology:graph_topology}

The design of the graph topology determines the information encapsulated by the graph; as such, it is one of the most important factors for the effectiveness of graph representation methods \cite{dong_learning_2019}. Given a set of transactions $X$, we want to create a graph $\mathcal{G}$, which captures the attribute information of the transactions within $X$ and their relations.

The aim of the Graph Construction Phase is to convert each daily batch of transactions $B_i$ into a graph representation $\mathcal{G}_{i|\theta}$ from which contrastive instance pairs can be sampled. Formally, for each transaction batch $B_i$ we construct a weigthed directed graph $\mathcal{G}_{i|\theta} = (\mathcal{V}, \mathcal{E}, \mathbf{X}^{\mathrm{feat}}, w)$, where $\mathcal{V}$ is the set of nodes, $\mathcal{E}^u$ is the set of edges, $\mathbf{X}^{\mathrm{feat}}$ is the node feature matrix, and $w$ provides edge weight mapping. 

Here, each node $v_i \in \mathcal{V}$ represents a transaction in $x_i \in X = B_t \cup X_{t|\theta}$. The sampling window $\theta$ is distinct from $\eta$ because they have different functions within the learning framework. While $\eta$ controls the size of the training set, $\theta$ ensures that the subgraphs sampled for transactions in $B_i$ include transaction nodes up to $\theta$ days earlier. This approach of defining two separate sliding windows is based on the approach taken by traditional feature aggregation methods where aggregations are computed over a consistently sized aggregation window, not limited by the size of the training set \cite{van_vlasselaer_apate_2015}.

The set of edges $\mathcal{E}$ is subsequently defined as: 

\begin{equation}
    \mathcal{E} = \{ (v_i, v_j) \in \mathcal{V} \times \mathcal{V} \mid i > j \wedge x_{i,r} = x_{j,r}, r \in \mathcal{R} \}.
\label{eq:homogeneous_transaction_graph_edges}
\end{equation}
\noindent
where edges are directed from a source node $v_i$ and to a destination node $v_j$ and $i > j$ implies that $v_i$ represents a more recent transaction than $v_j$. Additionally, variables $x_{i,r}$ and $x_{j,r}$ denote the values of a categorical variable $r \in \mathcal{R}$ corresponding to the $i^{\mathrm{th}}$ and $j^{\mathrm{th}}$ transactions, respectively. Finally, $\mathcal{R}$ represents a subset of the categorical variables within $X$. Note that by this description, the set of edges $\mathcal{E}$ is a multiset, making $\mathcal{G}$ a multigraph where multiple parallel edges can exist between a pair of transaction nodes given that they share the same value for at least two categorical features $r, r' \in \mathcal{R}$ and $r \neq r'$.

The categorical nature of credit card transaction features drives the construction of edges in our framework. Typically, categorical features are converted into numerical values \cite{van_vlasselaer_apate_2015}, but these features also indicate relationships among transactions sharing the same category. For example, transactions with the same CardID or MerchantID likely exhibit similar characteristics. 

Additionally, directed edges from recent to older transactions ensure that each target node $v_j$'s subgraph only includes preceding transactions $v_j'$ (where $j>j'$). This design mirrors real-world Fraud Detection Systems (FDS) requirements, where the model must assess transaction anomalies in near real-time, without future data. This constraint is maintained during training, teaching the model to detect anomalies without future information.

The set of node features is a column-wise slice of $\mathbf{X}$, where $\mathbf{X}$ is the matrix representation of the feature vectors within $X$, such that $\mathbf{X}^{\mathrm{feat}} = (\mathbf{X}_{*,f})_{ f \in \mathcal{F}}$. Here $\mathcal{F}$ is a subset of the features within $X$.

We define an edge weight function $w: \mathcal{E}_t \rightarrow \mathbb{R}$ which maps edges to their corresponding edge weight. Given the dynamic nature of cardholder spending and fraud vectors, the edge weights inverse the time elapsed between the two transaction nodes connected by an edge. This means that the edge weights between transactions will be determined by how close the two transactions occur in time, with transactions occurring closer in time being connected by edges with higher weights reflecting a stronger relationship. Formally the function $w$ is defined as: 

\begin{equation}
    w(e_{i,j}) =  t_{range} - (x_{i,\mathrm{time}} - x_{j,\mathrm{time}}),
\label{eq:transaction_graph_weight_function}
\end{equation}
\noindent 
where $x_{i,\mathrm{time}}$ and $x_{j,\mathrm{time}}$ are the timestamps, measured in seconds, of transactions $x_i$ and $x_j$, respectively. Here, the $t_{range}$ is the time range between the oldest and the most recent transaction in $X$.

Lastly, the reason for defining a separate dynamic graph instance for each batch is twofold. Firstly, it allows us to keep the sampling window consistent for all transaction batches across the dataset partitions. Secondly, it allows for efficient memory usage as the instance pairs can be sampled from each dynamic graph instance at a time during preprocessing. Due to the high edge density of the graphs, the limiting factor for increasing the training dataset size $\eta$ or sampling window size $\theta$ is the size of the full graphs used during instance pair sampling. 

\subsection{Instance Pair Sampling}\label{sec:methodology:graph_topology}

The instance pair sampling phase of the framework is based on the CoLA framework by \cite{liu_anomaly_2022_COLA}. The concept of a contrastive instance pair in the CoLA framework is based on the idea that a node's anomaly often correlates with its relationship to neighbouring structures. By contrasting a single target node with its adjacent substructure, the framework aims to detect anomalies within attributed graphs. The instance pair sampling procedure involves four steps: target node selection, subgraph sampling, anonymization, and subgraph augmentation. 

\textbf{Target Node Selection}: Given a graph $\mathcal{G}_{i|\theta}$, this step is executed by selecting each node representing a transaction in $B_i$ as the target in a random sequence during an epoch, constituting stochastic sampling without replacement \cite{liu_anomaly_2022_COLA}. Note that this does not include nodes representing transactions in $X_{t|\theta}$ as these only serve to provide a historical context of spending patterns.

\textbf{Subgraph Sampling}: Sampling for positive and negative pairs involves designating the initial node as the target for the former and a randomly chosen node for the latter. Following the approach in~\cite{liu_anomaly_2022_COLA}, random walks with restarts are employed for local subgraph sampling, chosen for its usability and efficiency. 

\textbf{Anonymization}: The initial node in each subgraph is anonymized by resetting its feature vector to a zero vector. Anonymization hinders the contrastive learning model from easily discerning the presence of target nodes within local subgraphs \cite{liu_anomaly_2022_COLA}. 

\textbf{Subgraph Augmentation}: This step is our contribution. Here, the sampled subgraph is converted to a multi-relational undirected graph, where the set of multi-relational edges can formally be defined as: 

\begin{equation}
    \mathcal{E}^{\mathcal{MR}} = \{ (v_i, r, v_j) \in \mathcal{V} \times \mathcal{R} \times \mathcal{V} \mid x_{i,r} = x_{j,r} \},
\label{eq:heterogeneous_transaction_graph_edges}
\end{equation}
\noindent
where $v_i$ and $v_j$ represent two nodes within the set $\mathcal{V}$ and $\mathcal{R}$ is a subset of the categorical features in $X$. The terms $x_{i,r}$ and $x_{j,r}$ denote the values of a specific categorical variable $r$ corresponding to the $i^{\text{th}}$ and $j^{\text{th}}$ transactions, respectively. Most importantly, each edge $(v_i, r, v_j) \in \mathcal{E}^{\mathcal{HG}}$ keeps track of the relation $r$ behind the edge. This allows additional information about the nature of transaction relatedness to be captured by the graph data structure. Additionally, the subgraph is converted to an undirected graph to facilitate effective message passing when applying the GNNs.

\subsection{GNN-based Contrastive Learning}\label{sec:methodology:contrastive_framework}

The GNN-based Contrastive Learning phase is also based on the CoLA framework \cite{liu_anomaly_2022_COLA}. This phase comprises three main modules: a GNN, a readout module, and a discriminator. 

\textbf{GNN Module}: The GNN is used to compute node embeddings of the nodes within the local subgraph and the isolated target node. In our experiments, we employ a GCN \cite{kipf_semi-supervised_2017} for uni-relational graphs and an R-GCN \cite{schlichtkrull_modeling_2018} for multi-relational graphs.

\textbf{Readout Module}: The readout module synthesizes the subgraph node embeddings into a single embedding vector. An average pooling readout function is employed here, like in \cite{liu_anomaly_2022_COLA}. 

\textbf{Discriminator Module}: The discriminator module processes the two embeddings to predict whether the target node belongs to the subgraph (i.e. whether the instance pair is a positive or negative pair). This module uses a bilinear scoring function, the same approach as in \cite{liu_anomaly_2022_COLA, zheng_generative_2021}.

\textbf{Anomaly Score Computation}: The goal is that the discriminator outputs a score of 1 for positive pairs and a score of 0 for negative pairs, reflecting the compatibility between nodes and their local substructures. Based on this \citet{liu_anomaly_2022_COLA} define an anomaly score function $f(v_i)$ as the difference between the negative and positive predicted scores of a node $v_i$:

\begin{equation}
f(v_i) = s_i^{(-)} - s_i^{(+)},
\label{eq:anomaly_score}
\end{equation}
\noindent
where $s_i^{(-)}$ and $s_i^{(+)}$ are the predicted scores for the negative instance pair and the positive instance pair, respectively. For normal nodes, the positive score $s^{(+)}$ should approximate 1, and the negative score $s^{(-)}$ 0. Both scores should converge around 0.5 for fraudulent nodes, reflecting their anomaly. It follows that the anomaly score of genuine nodes should converge towards -1, while for fraudulent nodes, it should settle around 0, allowing for effective discrimination of the anomalous fraud nodes.

To enhance anomaly detection, \citet{liu_anomaly_2022_COLA} introduce a multi-round, positive-negative sampling approach. This involves processing $R$ positive and $R$ negative instance pairs for each node $v_i$. The node's anomaly score is the average of score differences across these $R$ rounds. This method captures the variance in a node's normality across its neighbourhood and is critical for identifying anomalies that might be overlooked in single-shot sampling.

\section{Experiments}

\subsection{Datasets}

Two datasets were utilized to assess our proposed methods: a real-world dataset from an industry partner and a synthetic dataset. The real-world dataset contains approximately 700,000 transactions. The original dataset had significantly lower fraud rates than other datasets in the literature. Thus, the set of cards containing no fraudulent transactions was downsampled to emulate an issuer with fewer cards but a higher fraud rate and only transaction types with significant fraud rates were included. As a result, the dataset has a fraud rate of 0.41\%, aligning with those in existing literature (e.g. \cite{van_belle_inductive_2022}), enabling cross-study comparison. The synthetic dataset, generated by the Sparkov Data Generation tool \cite{harris_generate_2023} and curated by \cite{grover_fdb_2022}, comprises 1.3 million transactions with a 0.83\% fraud rate. It was selected for its necessary features (e.g., merchant ID, card ID) that facilitate graph construction with topology comparable to the private dataset.

A rolling window methodology was implemented to generate successive splits of the validation, training and test datasets, a widely adopted approach in the literature (e.g. \cite{dal_pozzolo_credit_2018}). This facilitates the replication of each experiment across different test days.

Our models only work with numerical data, so categorical attributes like country, currency, and merchant category codes (MCC) are converted into numerical formats. While common, one-hot encoding can lead to attribute dimension inflation, especially for categories with numerous unique values. An alternative is numerical mapping based on a fraud exposure score calculated from each category's fraud ratio. This method aligns with contrastive learning by distinguishing transactions through categorical fraud risk profiles, reflecting research on varied fraud levels across categories \cite{van_vlasselaer_apate_2015}. This study uses risk-based encoding for the real-world dataset due to its high-cardinality categorical variables. We calculate risk scores excluding test period data to prevent information leakage, assigning a default risk score of 1 to test-exclusive categories, emphasizing their abnormality. Additionally, we normalize all numerical attributes to stabilize training.

\begin{table*}
    \centering
    \fontsize{10}{12}\selectfont
    \caption{Results on the real-world dataset - Averaged over 5 test days}
    \renewcommand{\arraystretch}{1.0} 
    \begin{tabular}{lcccrrr}
    \toprule
        & \multicolumn{3}{c}{\textbf{Method}} & \multicolumn{3}{c}{\textbf{Performance Metrics}} \\
        \cmidrule(lr){2-4} \cmidrule(lr){5-7}
        \textbf{Graph Relations ($\mathcal{R}$)} & \textbf{GG} & \textbf{WS} & \textbf{MR} & \textbf{PR-AUC $\uparrow$} & \textbf{F1-Score $\uparrow$} & \textbf{NPr@100 $\uparrow$} \\

        \midrule
       
        \multirow{4}{*}{\textbf{Card ID}} & \checkmark & & & \textbf{9.78} \specialcell{± \scriptsize 4.31} & \textbf{15.62} \specialcell{± \scriptsize 5.53} & \textbf{22.72} \specialcell{± \scriptsize 3.87} \\ 
        & \checkmark & \checkmark & & 7.38 \specialcell{± \scriptsize 2.83} & 14.69 \specialcell{± \scriptsize 2.89} & 20.27 \specialcell{± \scriptsize 2.27} \\
        & \checkmark & & \checkmark & 2.64 \specialcell{± \scriptsize 1.12} & 7.33 \specialcell{± \scriptsize 3.45} & 9.25 \specialcell{± \scriptsize 3.98} \\
        & \checkmark & \checkmark & \checkmark & 0.92 \specialcell{± \scriptsize 0.28} & 2.45 \specialcell{± \scriptsize 1.41} & 2.75 \specialcell{± \scriptsize 1.81} \\

        \midrule
       
        \multirow{4}{*}{\textbf{Card ID + Merchant ID}} & \checkmark & & & 2.19 \specialcell{± \scriptsize 1.28} & 5.46 \specialcell{± \scriptsize 3.01} & 7.46 \specialcell{± \scriptsize 3.66} \\
        & \checkmark & \checkmark & & 2.38 \specialcell{± \scriptsize 1.17} & 5.90 \specialcell{± \scriptsize 4.50} & 7.90 \specialcell{± \scriptsize 3.42} \\
        & \checkmark & & \checkmark & \textbf{3.15} \specialcell{± \scriptsize 1.34} & \textbf{8.62} \specialcell{± \scriptsize 3.48} & \textbf{11.41} \specialcell{± \scriptsize 4.35} \\
        & \checkmark & \checkmark & \checkmark & 3.02 \specialcell{± \scriptsize 1.11} & 8.18 \specialcell{± \scriptsize 2.59} & 11.24 \specialcell{± \scriptsize 4.29} \\

    \bottomrule
    \multicolumn{7}{l}{\scalebox{0.7}{* Performance metric scores are presented in \% incl. standard deviations over five experiments initiated with different random seeds.}}
    \end{tabular}
    \label{tab:real_world_results}
\end{table*}

\subsection{Evaluation Metrics}
The approaches are evaluated based on three performance metrics: two are generally suited to evaluating classification performance in highly imbalanced datasets, and one is a domain-specific metric.

The first metric is the \textit{area under the precision-recall curve} (PR-AUC). The PR-AUC metric is preferential to the Area Under the ROC curve when working with highly imbalanced datasets~\cite{davis_relationship_2006}. We also evaluate the models using their \textit{F1-score} to provide a threshold-dependent metric. The F1-score represents the harmonic mean of precision and recall. Due to its threshold dependence, the threshold is optimized using the validation set.

Finally, the number of alerts that investigators can check daily is limited due to resource constraints. This means that practitioners are highly concerned about the precision of the limited amount of transactions that can be alerted \cite{dal_pozzolo_credit_2018}. This is captured by the \textit{alert precision} statistic, $Pr@k = \frac{|TP_k|}{k}$, which measures the precision among the $k$ alerted transactions, where $|TP_k|$ denotes the number of true positives within alerts the $k$ most risky transactions. To adjust for the days where the number of frauds is less than $k$, \cite{dal_pozzolo_credit_2018} introduce a \textit{normalized alert precision}: 

\begin{equation}
    NPr@k = \frac{Pr@k}{\Gamma}, \Gamma = \begin{cases}
                                            1 \textrm{ if } |F_t| \ge k\\
                                            \frac{|F_t|}{k} \textrm{ if } |F_t| < k
                                            \end{cases}
    \label{eq:Pr@k}
\end{equation}{}
\noindent
where $\Gamma$ is a normalization factor and $|F_t|$ is the number of frauds on day $t$.

\subsection{Baselines}
This study evaluates the plain GraphGuard (GG) setup against set-ups where weighted sampling (WS) and the multi-relational (MR) subgraph augmentation are added incrementally, resulting in four distinct approaches.

Additionally, two graph definitions are used. The first definition has edges between transactions that share a Card ID or a Merchant ID (i.e. $\mathcal{R} = [\mathrm{CardID}, \mathrm{MerchantID}]$). In contrast, the second definition only has Card ID edges (i.e. $\mathcal{R} = [\mathrm{CardID}]$). These two graph definitions are motivated by traditional feature aggregation methods. These approaches typically aggregate features on both the card and merchant level, suiting to highlight abnormal transactions based on patterns on the same card or at the same merchant, respectively \cite{van_vlasselaer_apate_2015}. Moreover, we also evaluate the approaches on graphs having only Card ID relations as feature importance studies suggest that card-level aggregation is a stronger discriminator for fraud detection. Furthermore, previous studies on contrastive graph-based approaches to anomaly detection have shown that these approaches perform better in graphs with lower mean degrees, as edges typically indicate a stronger relationship \cite{zheng_generative_2021, liu_anomaly_2022_COLA}. As, on average, each merchant has more transactions associated with it than each card, there will be significantly more merchant edges than card edges. 

\subsection{Parameter Settings}

Model parameters were optimised on graphs with both Card ID and Merchant ID relations through a grid search. For all methods, a subgraph size of 2 was consistently used. The Graph Convolutional Networks (GCNs) and R-GCNs employed a single-layer GNN architecture. This design choice follows the rationale presented in \cite{liu_anomaly_2022_COLA}, asserting its adequacy for computing representations in small subgraphs. For the real-world dataset, all models were trained for 40 epochs with a learning rate of 0.0001, a batch size of 1024, and an embedding size of 8. On the synthetic dataset, the models were trained for 100 epochs with a learning rate of 0.00001. Here, the GCNs were trained using a batch size of 1024 and an embedding size of 8. In contrast, the R-GCNs were trained with a smaller batch size of 128 and an embedding size of 2. Finally, due to the significantly lower density of the graphs with Card ID-only relations, these graphs have an aggregation window of $\theta= 30$ days, while the graphs with both Card ID and Merchant ID relations only have an aggregation window of $\theta= 7$ days.

\begin{table*}
    \centering
    \fontsize{10}{12}\selectfont
    \caption{Results on the synthetic dataset - Averaged over 9 test days}
    \renewcommand{\arraystretch}{1.0} 
    \begin{tabular}{lcccrrr}
    \toprule
        & \multicolumn{3}{c}{\textbf{Method}} & \multicolumn{3}{c}{\textbf{Performance Metrics}} \\
        \cmidrule(lr){2-4} \cmidrule(lr){5-7}
        \textbf{Graph Relations ($\mathcal{R}$)} & \textbf{GG} & \textbf{WS} & \textbf{MR} & \textbf{PR-AUC $\uparrow$} & \textbf{F1-Score $\uparrow$} & \textbf{NPr@100 $\uparrow$} \\
        \midrule
       
        \multirow{4}{*}{\textbf{Card ID}} & \checkmark & & & 12.81 \specialcell{± \scriptsize 14.42} & 16.17 \specialcell{± \scriptsize 21.67} & \textbf{27.91} \specialcell{± \scriptsize 18.31} \\ 
        & \checkmark & \checkmark & & 11.10 \specialcell{± \scriptsize 19.31} & 13.17 \specialcell{± \scriptsize 18.88} & 21.41 \specialcell{± \scriptsize 26.14} \\
        & \checkmark & & \checkmark & 12.31 \specialcell{± \scriptsize 18.31} & 16.76 \specialcell{± \scriptsize 23.12} & 22.95 \specialcell{± \scriptsize 24.59} \\
        & \checkmark & \checkmark & \checkmark & \textbf{14.98} \specialcell{± \scriptsize 16.95} & \textbf{17.47} \specialcell{± \scriptsize 19.79} & 26.44 \specialcell{± \scriptsize 25.29} \\

        \midrule
       
        \multirow{4}{*}{\textbf{Card ID \& Merchant ID}} & \checkmark & & & \textbf{6.80} \specialcell{± \scriptsize 7.58}  & \textbf{2.31} \specialcell{± \scriptsize 1.31} & 7.85 \specialcell{± \scriptsize 8.81} \\
        & \checkmark & \checkmark &  & 5.93 \specialcell{± \scriptsize 7.81} & 2.08 \specialcell{± \scriptsize 1.15} & \textbf{13.34} \specialcell{± \scriptsize 15.00} \\
        & \checkmark &  & \checkmark & 1.90 \specialcell{± \scriptsize 2.87} & 1.64 \specialcell{± \scriptsize 1.68} & 7.07 \specialcell{± \scriptsize 7.81} \\
        & \checkmark & \checkmark & \checkmark & 2.09 \specialcell{± \scriptsize 3.39} & 1.42 \specialcell{± \scriptsize 1.77} & 7.18 \specialcell{± \scriptsize 12.30} \\

    \bottomrule
    \multicolumn{7}{l}{\scalebox{0.7}{* Performance metric scores are presented in \% incl. standard deviations over nine experiments initiated with different random seeds.}}
    \end{tabular}
    \label{tab:synthetic_results}
\end{table*}

\subsection{Results}

This section presents the preliminary results of the proposed methods tested on both real-world and synthetic datasets. Each method's performance is averaged over multiple test days and model initialization seeds, with standard deviations provided. 

Table \ref{tab:real_world_results} presents the results averaged over 5 test days on the real-world dataset. Firstly, the uni-relational methods perform significantly better on the graphs with only the Card ID relations. This could be due to the lower average degree within these graphs, which would align with the findings of \cite{zheng_generative_2021, liu_anomaly_2022_COLA}, who find similar contrastive approaches to perform better in graphs with lower mean degrees. The reason is that nodes with fewer edges are more likely to be more tightly related to the nodes they are connected to. Additionally, previous studies on feature aggregation approaches in the credit card fraud detection literature indicate that card-level spending patterns provide a stronger discriminative power in terms of identifying fraudulent transactions than merchant-level spending patterns \cite{van_vlasselaer_apate_2015}. This could be because merchant-level spending patterns might be easier for fraudsters to imitate, as fraudsters are less likely to be intimately familiar with the spending patterns of the cardholder they are defrauding. However, these graphs also have a larger aggregation window (30 days vs. 7 days) than the graphs with both Card ID and Merchant ID edges, which could also be a reason behind the performance improvements. Additional experiments are needed to tease out the individual effects of these two potential sources of improvement.

Secondly, the performance on the uni-relational graphs is significantly higher than the performance on the multi-relational graphs with only the Card ID edges. In contrast, the multi-relational approach outperforms the uni-relational approach on the graph with the two edge relations. On the one hand, the improved performance of the multi-relational approach makes sense on multi-relational graphs as the R-GCN is able to treat these relations differently through the use of separate weight matrices. On the other hand, the significant decrease in performance on the graphs with only Card ID relations is surprising. One would not expect the R-GCN to perform so much worse than the GCN here. One potential reason could be that the model hyperparameters were only tuned on the graphs with both Card ID and Merchant ID relations. Thus, it is conceivable that the optimal hyperparameters for the GCN translate better between the two different graph topologies. Moreover, unlike the GCN, the R-GCN has a dedicated weight matrix for the self-loop aggregation. Thus, another explanation for this performance divergence could be that the GCN's shared weights and, thus, lower complexity allow it to converge better within the training epochs.  

Finally, the results in Table \ref{tab:real_world_results} indicate that adding weighted sampling hinders performance, with only one out of four methods showing improved performance. The reason for this could be that fraudulent transactions typically occur closely clustered in time as fraudsters usually attempt many transactions in a relatively short amount of time in order to empty the card before the fraud is discovered and the card is blocked \cite{sahin2013cost}. The time-weighted sampling might influence the sampling process to sample other fraudulent transactions within the local subgraph of fraudulent transactions occurring later in this fraud sequence, thus allowing them to camouflage from the contrastive model.





Table \ref{tab:synthetic_results} presents the results obtained on the synthetic dataset, averaged over 9 test days. Similar to the previously discussed results, the best performance is achieved on the graphs with only the Card ID relations. This further suggests that either the subgraphs sampled in this case offer a more discriminative contrast or that the larger aggregation window provides improved information on previous spending patterns. However, here, the multi-relational approach performs better in the Card ID only graphs, which could be because the hyperparameters tuned on the other graph might transform better there than on the real-world dataset by chance. Lastly, the weighted sampling improves performance here in most cases. This could be due to the fact that this synthetic dataset does not emulate the clustered nature of fraudulent transactions apparent in real-world datasets. 

\section{Conclusion and Future Work}

In this study, we presented GraphGuard. We developed a novel graph representation for transaction networks and adapted a state-of-the-art self-supervised anomaly detection framework to this domain. The framework was evaluated on two datasets, including a real-world one.

Results show that self-supervised graph representation learning effectively detects credit card fraud without labelled data. Temporal relationships between transactions, incorporated via time-weighted subgraph sampling, do not seem to enhance performance. Moreover, the benefit of relational information varied across datasets and graph topologies.

Future research will integrate a supervision signal for model fine-tuning and improve relational information processing. \cite{van_belle_catchm_2023} emphasized the value of supervision in learning effective node representations for fraud detection, and \cite{zheng_unsupervised_2022} highlighted the utility of limited labelled data in self-supervised learning for graph-based anomaly detection. Further, exploring the use of Card ID edges for sampling contrastive pairs and incorporating multi-relational information during the graph representation learning phase is a promising area for future investigation. This would involve using one set of edges for subgraph sampling while augmenting subgraphs with diverse relations for richer representations.

\bibliography{aaai24}

\begin{thebibliography}{32}
\providecommand{\natexlab}[1]{#1}

\bibitem[{Akoglu, Tong, and Koutra(2015)}]{akoglu_graph_2015}
Akoglu, L.; Tong, H.; and Koutra, D. 2015.
\newblock Graph based anomaly detection and description: a survey.
\newblock \emph{Data Mining and Knowledge Discovery}, 29(3): 626--688.

\bibitem[{Alippi, Boracchi, and Roveri(2013)}]{alippi2013just}
Alippi, C.; Boracchi, G.; and Roveri, M. 2013.
\newblock Just-in-time classifiers for recurrent concepts.
\newblock \emph{IEEE transactions on neural networks and learning systems}, 24(4): 620--634.

\bibitem[{Bronstein et~al.(2017)Bronstein, Bruna, LeCun, Szlam, and Vandergheynst}]{bronstein_geometric_2017}
Bronstein, M.~M.; Bruna, J.; LeCun, Y.; Szlam, A.; and Vandergheynst, P. 2017.
\newblock Geometric {Deep} {Learning}: {Going} beyond {Euclidean} data.
\newblock \emph{IEEE Signal Processing Magazine}, 34(4): 18--42.
\newblock Conference Name: IEEE Signal Processing Magazine.

\bibitem[{Button, Lewis, and Tapley(2014)}]{button_not_2014}
Button, M.; Lewis, C.; and Tapley, J. 2014.
\newblock Not a victimless crime: {The} impact of fraud on individual victims and their families.
\newblock \emph{Security Journal}, 27(1): 36--54.

\bibitem[{Carcillo et~al.(2017)Carcillo, Dal~Pozzolo, Le~Borgne, Caelen, Mazzer, and Bontempi}]{carcillo_scarff_2017}
Carcillo, F.; Dal~Pozzolo, A.; Le~Borgne, Y.-A.; Caelen, O.; Mazzer, Y.; and Bontempi, G. 2017.
\newblock {SCARFF} : a {Scalable} {Framework} for {Streaming} {Credit} {Card} {Fraud} {Detection} with {Spark}.
\newblock \emph{Information Fusion}, 41.

\bibitem[{Dal~Pozzolo et~al.(2018)Dal~Pozzolo, Boracchi, Caelen, Alippi, and Bontempi}]{dal_pozzolo_credit_2018}
Dal~Pozzolo, A.; Boracchi, G.; Caelen, O.; Alippi, C.; and Bontempi, G. 2018.
\newblock Credit {Card} {Fraud} {Detection}: {A} {Realistic} {Modeling} and a {Novel} {Learning} {Strategy}.
\newblock \emph{IEEE Transactions on Neural Networks and Learning Systems}, 29(8): 3784--3797.
\newblock Conference Name: IEEE Transactions on Neural Networks and Learning Systems.

\bibitem[{Dal~Pozzolo et~al.(2015)Dal~Pozzolo, Caelen, Johnson, and Bontempi}]{dal2015calibrating}
Dal~Pozzolo, A.; Caelen, O.; Johnson, R.~A.; and Bontempi, G. 2015.
\newblock Calibrating probability with undersampling for unbalanced classification.
\newblock In \emph{2015 IEEE symposium series on computational intelligence}, 159--166. IEEE.

\bibitem[{Davis and Goadrich(2006)}]{davis_relationship_2006}
Davis, J.; and Goadrich, M. 2006.
\newblock The relationship between {Precision}-{Recall} and {ROC} curves.
\newblock In \emph{Proceedings of the 23rd international conference on {Machine} learning}, {ICML} '06, 233--240. New York, NY, USA: Association for Computing Machinery.
\newblock ISBN 978-1-59593-383-6.

\bibitem[{Dong et~al.(2019)Dong, Thanou, Rabbat, and Frossard}]{dong_learning_2019}
Dong, X.; Thanou, D.; Rabbat, M.; and Frossard, P. 2019.
\newblock Learning graphs from data: {A} signal representation perspective.
\newblock \emph{IEEE Signal Processing Magazine}, 36(3): 44--63.
\newblock ArXiv:1806.00848 [cs, stat].

\bibitem[{Grover et~al.(2022)Grover, Li, Liu, Zablocki, Zhou, Xu, and Cheng}]{grover_fdb_2022}
Grover, P.; Li, Z.; Liu, J.; Zablocki, J.; Zhou, H.; Xu, J.; and Cheng, A. 2022.
\newblock {FDB}: {Fraud} {Dataset} {Benchmark}.
\newblock ArXiv:2208.14417 [cs, stat].

\bibitem[{Harris(2023)}]{harris_generate_2023}
Harris, B. 2023.
\newblock Generate {Fake} {Credit} {Card} {Transaction} {Data}, {Including} {Fraudulent} {Transactions}.
\newblock Original-date: 2016-02-09T15:19:00Z.

\bibitem[{Hojjati, Ho, and Armanfard(2023)}]{hojjati_self_supervised_anomaly_survey_2023}
Hojjati, H.; Ho, T. K.~K.; and Armanfard, N. 2023.
\newblock Self-{Supervised} {Anomaly} {Detection}: {A} {Survey} and {Outlook}.
\newblock ArXiv:2205.05173 [cs].

\bibitem[{Kipf and Welling(2017)}]{kipf_semi-supervised_2017}
Kipf, T.~N.; and Welling, M. 2017.
\newblock Semi-{Supervised} {Classification} with {Graph} {Convolutional} {Networks}.
\newblock Toulon, France: arXiv.
\newblock ArXiv:1609.02907 [cs, stat].

\bibitem[{Lebichot et~al.(2017)Lebichot, Braun, Caelen, and Saerens}]{lebichot_graph_semi_2017}
Lebichot, B.; Braun, F.; Caelen, O.; and Saerens, M. 2017.
\newblock A graph-based, semi-supervised, credit card fraud detection system.
\newblock In Cherifi, H.; Gaito, S.; Quattrociocchi, W.; and Sala, A., eds., \emph{Complex {Networks} \& {Their} {Applications} {V}}, Studies in {Computational} {Intelligence}, 721--733. Cham: Springer International Publishing.
\newblock ISBN 978-3-319-50901-3.

\bibitem[{Liu et~al.(2021{\natexlab{a}})Liu, Tang, Tian, and Wang}]{liu_gnn_ccfraud_2021}
Liu, G.; Tang, J.; Tian, Y.; and Wang, J. 2021{\natexlab{a}}.
\newblock Graph {Neural} {Network} for {Credit} {Card} {Fraud} {Detection}.
\newblock In \emph{2021 {International} {Conference} on {Cyber}-{Physical} {Social} {Intelligence} ({ICCSI})}, 1--6. Beijing, China: IEEE.
\newblock ISBN 978-1-66542-621-3.

\bibitem[{Liu et~al.(2022{\natexlab{a}})Liu, HaoChen, Gaidon, and Ma}]{liu_SSL_imbalance_2022}
Liu, H.; HaoChen, J.~Z.; Gaidon, A.; and Ma, T. 2022{\natexlab{a}}.
\newblock Self-supervised {Learning} is {More} {Robust} to {Dataset} {Imbalance}.
\newblock ArXiv:2110.05025 [cs, stat].

\bibitem[{Liu et~al.(2023)Liu, Zhang, Hou, Mian, Wang, Zhang, and Tang}]{liu_self_supervised_survey_2023}
Liu, X.; Zhang, F.; Hou, Z.; Mian, L.; Wang, Z.; Zhang, J.; and Tang, J. 2023.
\newblock Self-{Supervised} {Learning}: {Generative} or {Contrastive}.
\newblock \emph{IEEE Transactions on Knowledge and Data Engineering}, 35(1): 857--876.
\newblock Conference Name: IEEE Transactions on Knowledge and Data Engineering.

\bibitem[{Liu et~al.(2022{\natexlab{b}})Liu, Li, Pan, Gong, Zhou, and Karypis}]{liu_anomaly_2022_COLA}
Liu, Y.; Li, Z.; Pan, S.; Gong, C.; Zhou, C.; and Karypis, G. 2022{\natexlab{b}}.
\newblock Anomaly {Detection} on {Attributed} {Networks} via {Contrastive} {Self}-{Supervised} {Learning}.
\newblock \emph{IEEE Transactions on Neural Networks and Learning Systems}, 33(6): 2378--2392.
\newblock Conference Name: IEEE Transactions on Neural Networks and Learning Systems.

\bibitem[{Liu et~al.(2021{\natexlab{b}})Liu, Pan, Wang, Xiong, Wang, Chen, and Lee}]{liu_anomaly_2021}
Liu, Y.; Pan, S.; Wang, Y.~G.; Xiong, F.; Wang, L.; Chen, Q.; and Lee, V.~C. 2021{\natexlab{b}}.
\newblock Anomaly {Detection} in {Dynamic} {Graphs} via {Transformer}.
\newblock \emph{IEEE Transactions on Knowledge and Data Engineering}, 1--1.
\newblock Conference Name: IEEE Transactions on Knowledge and Data Engineering.

\bibitem[{Ma et~al.(2021)Ma, Wu, Xue, Yang, Zhou, Sheng, Xiong, and Akoglu}]{ma_comprehensive_2021}
Ma, X.; Wu, J.; Xue, S.; Yang, J.; Zhou, C.; Sheng, Q.~Z.; Xiong, H.; and Akoglu, L. 2021.
\newblock A {Comprehensive} {Survey} on {Graph} {Anomaly} {Detection} with {Deep} {Learning}.
\newblock \emph{IEEE Transactions on Knowledge and Data Engineering}, 1--1.
\newblock Conference Name: IEEE Transactions on Knowledge and Data Engineering.

\bibitem[{Maynard and Thok(2022)}]{maynard_fighting_2022}
Maynard, N.; and Thok, H. 2022.
\newblock Fighting {Online} {Payment} {Fraud} in 2022 \& {Beyond}.
\newblock Whitepaper, Juniper Research Ltd.

\bibitem[{Paldino et~al.(2022)Paldino, Lebichot, Le~Borgne, Siblini, Oblé, Boracchi, and Bontempi}]{paldino_role_2022}
Paldino, G.~M.; Lebichot, B.; Le~Borgne, Y.-A.; Siblini, W.; Oblé, F.; Boracchi, G.; and Bontempi, G. 2022.
\newblock The role of diversity and ensemble learning in credit card fraud detection.
\newblock \emph{Advances in Data Analysis and Classification}.

\bibitem[{Ren et~al.(2015)Ren, He, Girshick, and Sun}]{ren_faster_2015}
Ren, S.; He, K.; Girshick, R.; and Sun, J. 2015.
\newblock Faster {R}-{CNN}: {Towards} {Real}-{Time} {Object} {Detection} with {Region} {Proposal} {Networks}.
\newblock In \emph{Advances in {Neural} {Information} {Processing} {Systems}}, volume~28. Curran Associates, Inc.

\bibitem[{Ryman-Tubb, Krause, and Garn(2018)}]{ryman-tubb_how_2018}
Ryman-Tubb, N.~F.; Krause, P.; and Garn, W. 2018.
\newblock How {Artificial} {Intelligence} and machine learning research impacts payment card fraud detection: {A} survey and industry benchmark.
\newblock \emph{Engineering Applications of Artificial Intelligence}, 76: 130--157.

\bibitem[{Sahin, Bulkan, and Duman(2013)}]{sahin2013cost}
Sahin, Y.; Bulkan, S.; and Duman, E. 2013.
\newblock A cost-sensitive decision tree approach for fraud detection.
\newblock \emph{Expert Systems with Applications}, 40(15): 5916--5923.

\bibitem[{Schlichtkrull et~al.(2018)Schlichtkrull, Kipf, Bloem, van den Berg, Titov, and Welling}]{schlichtkrull_modeling_2018}
Schlichtkrull, M.; Kipf, T.~N.; Bloem, P.; van den Berg, R.; Titov, I.; and Welling, M. 2018.
\newblock Modeling {Relational} {Data} with {Graph} {Convolutional} {Networks}.
\newblock In Gangemi, A.; Navigli, R.; Vidal, M.-E.; Hitzler, P.; Troncy, R.; Hollink, L.; Tordai, A.; and Alam, M., eds., \emph{The {Semantic} {Web}}, Lecture {Notes} in {Computer} {Science}, 593--607. Cham: Springer International Publishing.
\newblock ISBN 978-3-319-93417-4.

\bibitem[{{UK Finance}(2022)}]{uk_finance_uk_2022}
{UK Finance}. 2022.
\newblock {UK} {Finance} calls for urgent action from all sectors as fraud continues to threaten the {UK}.

\bibitem[{Van~Belle, Baesens, and De~Weerdt(2023)}]{van_belle_catchm_2023}
Van~Belle, R.; Baesens, B.; and De~Weerdt, J. 2023.
\newblock {CATCHM}: {A} novel network-based credit card fraud detection method using node representation learning.
\newblock \emph{Decision Support Systems}, 164: 113866.

\bibitem[{Van~Belle et~al.(2022)Van~Belle, Van~Damme, Tytgat, and De~Weerdt}]{van_belle_inductive_2022}
Van~Belle, R.; Van~Damme, C.; Tytgat, H.; and De~Weerdt, J. 2022.
\newblock Inductive {Graph} {Representation} {Learning} for fraud detection.
\newblock \emph{Expert Systems with Applications}, 193: 116463.

\bibitem[{Van~Vlasselaer et~al.(2015)Van~Vlasselaer, Bravo, Caelen, Eliassi-Rad, Akoglu, Snoeck, and Baesens}]{van_vlasselaer_apate_2015}
Van~Vlasselaer, V.; Bravo, C.; Caelen, O.; Eliassi-Rad, T.; Akoglu, L.; Snoeck, M.; and Baesens, B. 2015.
\newblock {APATE}: {A} novel approach for automated credit card transaction fraud detection using network-based extensions.
\newblock \emph{Decision Support Systems}, 75: 38--48.

\bibitem[{Zheng et~al.(2021)Zheng, Jin, Liu, Chi, Phan, and Chen}]{zheng_generative_2021}
Zheng, Y.; Jin, M.; Liu, Y.; Chi, L.; Phan, K.~T.; and Chen, Y.-P.~P. 2021.
\newblock Generative and {Contrastive} {Self}-{Supervised} {Learning} for {Graph} {Anomaly} {Detection}.
\newblock \emph{IEEE Transactions on Knowledge and Data Engineering}, 1--1.
\newblock Conference Name: IEEE Transactions on Knowledge and Data Engineering.

\bibitem[{Zheng et~al.(2022)Zheng, Jin, Liu, Chi, Phan, Pan, and Chen}]{zheng_unsupervised_2022}
Zheng, Y.; Jin, M.; Liu, Y.; Chi, L.; Phan, K.~T.; Pan, S.; and Chen, Y.-P.~P. 2022.
\newblock From {Unsupervised} to {Few}-shot {Graph} {Anomaly} {Detection}: {A} {Multi}-scale {Contrastive} {Learning} {Approach}.
\newblock ArXiv:2202.05525 [cs].

\end{thebibliography}

\end{document}